\pdfoutput=1

\documentclass[journal]{IEEEtran}

\usepackage{cite}

\ifCLASSINFOpdf
  \usepackage[pdftex]{graphicx}
  \graphicspath{{./}}
  \DeclareGraphicsExtensions{.pdf,.jpeg,.jpg,.png,.eps}

\else
\fi
\usepackage{amsmath,amssymb}
\newcommand{\matr}[1]{#1}

\renewcommand{\vec}[1]{\mathbf{#1}}

\usepackage{stmaryrd}

\usepackage{caption} 
\usepackage{balance}

\begin{document}
%
\title{Training Passive Photonic Reservoirs with Integrated Optical Readout}
%
%
%

\author{Matthias~Freiberger,~\IEEEmembership{Student Member,~IEEE,}
                Andrew~Katumba,~\IEEEmembership{Student Member,~IEEE,}
               Peter~Bienstman,~\IEEEmembership{Member,~IEEE,}
        and~Joni~Dambre,~\IEEEmembership{Member,~IEEE}

\thanks{M.\ Freiberger and J.\ Dambre are with the Ghent University - imec IDLab, Department of Electronics and Information Systems, Technologiepark-Zwijnaarde 15, B-9052 Ghent, Belgium. A.\ Katumba and P.\ Bienstman are with the Ghent University - imec Photonics Research Group, Department of Information Technology, Technologiepark-Zwijnaarde 15, B-9052 Ghent, Belgium. \protect \\
Email: Matthias.Freiberger@UGent.be}
\thanks{Manuscript received 13.10.2017; revised 29.06.2018}}

\markboth{Transactions on Neural Networks and Learning Systems,~Vol.~X, No.~X, X~201X}%
{Freiberger \MakeLowercase{\textit{et al.}}: Training Passive Photonic Reservoirs with Integrated Optical Readout}

\IEEEpubid{TNNLS-2017-P-8539.R1~\copyright~2018 IEEE}

\maketitle


\begin{abstract}
As Moore's law comes to an end, neuromorphic approaches to computing are on the rise. One of these, passive photonic reservoir computing, is a strong candidate for computing at high bitrates (\textgreater 10 Gbps) and with low energy consumption. Currently though, both benefits are limited by the necessity to perform training and readout operations in the electrical domain. Thus, efforts are currently underway in the photonic community to design an integrated optical readout, which allows to perform all operations in the optical domain. In addition to the technological challenge of designing such a readout, new algorithms have to be designed in order to train it. Foremost, suitable algorithms need to be able to deal with the fact that the actual on-chip reservoir states are not directly observable. In this work, we investigate several options for such a training algorithm and propose a solution in which the complex states of the reservoir can be observed by appropriately setting the readout weights, while iterating over a predefined input sequence. We perform numerical simulations in order to compare our method with an ideal baseline requiring full observability as well as with an established black-box optimization approach (CMA-ES).  
\end{abstract}

\begin{IEEEkeywords}
Cognitive Computing, Reservoir Computing, Photonic Computing, Neuromorphic Computing, Nonlinearity Inversion, Integrated Optical Readout, Limited Observability.
\end{IEEEkeywords}

\IEEEpeerreviewmaketitle

\section{Introduction}
\label{sec:introduction}

\IEEEPARstart{A}{s} Moore's law appears to come to an end \cite{Haensch2006}, novel computing approaches, that go beyond traditional computing paradigms of silicon semiconductors are on the rise.  
Among these, Reservoir Computing \cite{Maas2002,Jaeger2004,Verstraeten2007}, a machine learning technique proposed to perform brain-inspired computing, has been applied in unconventional computing recently in attempts to overcome the Von Neumann bottleneck. While promising results have been shown on various hardware platforms \cite{Buerger2013,Demis2015,hermans2015photonic}, Photonic Reservoir Computing \cite{van2017advance,larger2012photonic,smerieri2012towards,fiers2014nanophotonic,vandoorne2011parallel, qin2017numerical} makes an especially strong case as a future-proof technology platform due to its potential to perform high-bandwidth tasks with little energy consumption. Especially, integrated passive photonic reservoir computing \cite{Vandoorne2014, Katumba2015, Katumba2017} is a promising approach since it exploits common CMOS fabrication technology and thus ensures an easy route to mass market production. Integrated passive photonic reservoirs have been shown to solve parity bit and header recognition tasks relevant for telecom applications at data rates higher than 10 Gbps.

\begin{figure}[!t]
\centering
\includegraphics[width=3.2in]{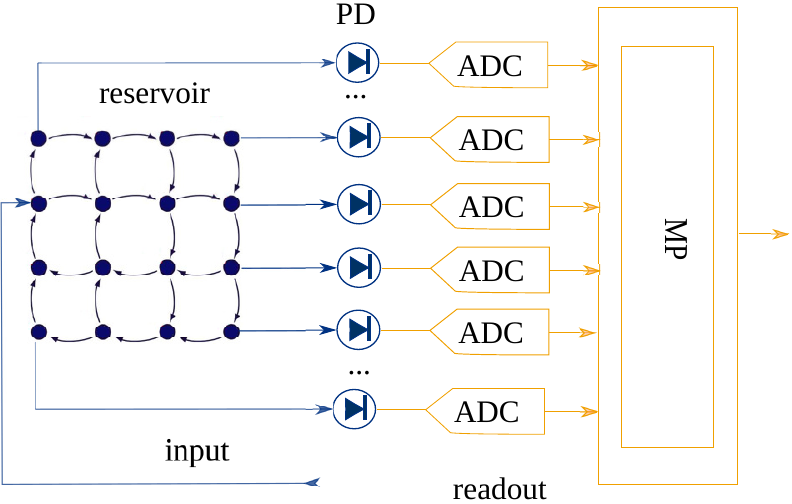}
\caption{Swirl reservoir and readout system of existing RC photonic chip prototypes, with the nodes of the reservoir being collected in the readout section, where optical output signals are converted to electrical signals and then processed to a final output. “PD”: photodiode, “ADC”: AD converter, “MP”: Microprocessor. The blue and orange parts represent respectively the optical and electronic signals and components.}
\label{fig:electrical_readout}
\end{figure}

Nevertheless, the technology is still afflicted by a number of limitations, which prevent its application to problems on a larger scale. One of these limitations lies in the fact that the training, as well as the mixing of signals to solve a meaningful task, has so far only happened in the electrical domain. Indeed, Vandoorne et al.\ \cite{Vandoorne2014} transferred the signal at each node from the optical to the electrical domain using a photodetector, and then sent it through an AD converter. Finally, the required linear combination of signals and readout weights was performed in the electrical domain using a microprocessor. See Figure \ref{fig:electrical_readout} for a detailed illustration of the process.

\begin{figure}[!t]
\centering
\includegraphics[width=3.2in]{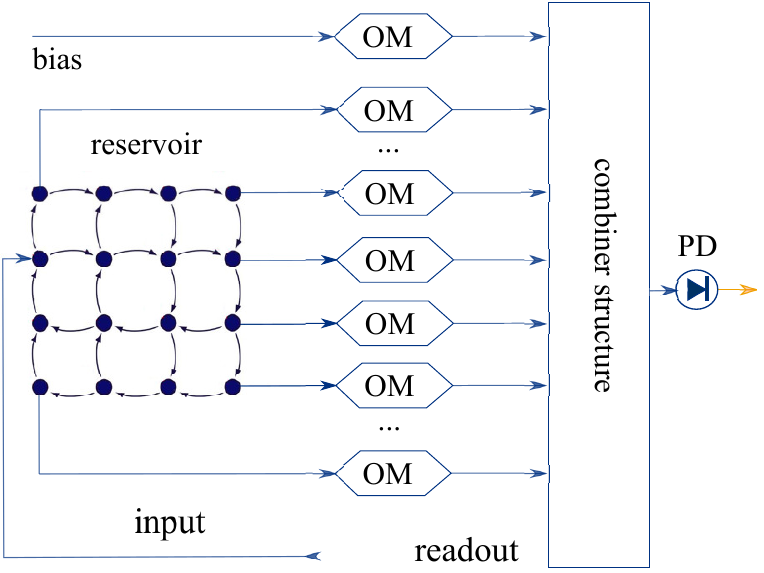}

\caption{Illustration of a swirl reservoir connected to a fully optical readout. Each optical output signal is modulated by an Optical Modulator (“OM”) implementing the weights. The optical outputs are then sent to a combiner structure where all signals are summed and finally converted to an electric output signal using a photo diode.}
\label{fig:optical_readout}
\end{figure}

\IEEEpubidadjcol 
In order to truly reap the benefits of optical computing though, signals need to be processed at very high data rates in an energy-efficient way. Considering ecological as well as economical factors, minimizing power consumption is of paramount importance for future computing technologies. From that perspective, the approach pursued in \cite{Vandoorne2014} for reading out integrated photonic reservoirs is inefficient, since there is a significant energy and latency cost associated to it. Hence, it is desirable to perform the summing of signals in the optical domain instead of in the electrical domain. 
Using such an integrated optical readout, only a single photodetector, which receives the weighted sum of all optical signals, is required. A straightforward low-power optical weighting element can take the form of a reverse-biased pn-junction. A better solution would be to use non-volatile optical weighting elements, such as the ones that are currently being developed by several groups \cite{Abel2013Astrong,rios2015integrated,van2015production}. The necessary summing of the individual weighted reservoir state signals can be performed by a combiner tree, i.e. by summing signals pairwise using an 2x1 optical combiner structure and repeating this step on the resulting intermediate signals until a single optical output signal is obtained. Figure \ref{fig:optical_readout} further illustrates the concept of a fully optical integrated readout.

However, by employing such an integrated optical readout, we lose direct observability of the states of the photonic reservoir. Observing all states is mandatory though, in order to use classical linear readout training algorithms such as ridge regression \cite{hoerl1970ridge} and other least-squares approaches. At first glance, a possible solution could be to add a separate high-speed photodetector to each reservoir node, which is only used during training to observe the states. The weights would then be calculated in the electrical domain, while the trained reservoir could still be operated entirely in the optical domain. This is unfortunately challenging due to a number of reasons. First, high-speed photodetectors tend to be costly in terms of chip footprint. Therefore, such an architecture would not scale well when increasing the number of nodes in the photonic reservoirs to numbers common in classic echo state networks \cite{Jaeger2004}. Second, since passive photonic reservoirs make use of coherent light for added richness, the tunable readout weights in the optical domain have to be complex-valued as well. However, unless we go to even more complicated coherent detectors, these photodiodes can only measure the intensities of the states and not their phases, so we cannot calculate the correct complex-valued weights. Note that this is different from the approach in \cite{Vandoorne2014}, which used real-valued weights on real-valued signals, whereas here we need to use complex-valued weights on complex-valued signals.

A second possible solution could be to train the weights based on simulations of the behaviour of a virtual reservoir, using photonic circuit simulation software, which will obviously have full observability of all the nodes. However, the fabrication tolerances of these devices are such that the propagation phase of two nominally identical waveguides could be completely different. This prohibits the successful transfer of weights trained using the idealized simulated reservoir to actual hardware. In summary: finding a suitable training algorithm is not straight-forward.

In this paper, we evaluate several approaches to train integrated photonic reservoirs without full state observability. We simulate the behavior of passive photonic reservoirs with integrated optical readout and train the simulated readouts of these reservoirs to perform 3 bit header recognition. We evaluate the investigated approaches by comparing their achieved bit error rates over a wide range of input signal bit rates. In Section \ref{sec:methodology}, we give an overview on our methodology and the simulation setup which is applied throughout the paper. Thereafter, in Section \ref{sec:complex-ridge}, we introduce a baseline approach inspired by classic reservoir training methods, against which all other approaches will be evaluated. We apply CMA-ES, a well known general-purpose black-box optimization approach, in Section \ref{sec:black-box-optimization} to train simulated reservoirs with integrated readout and evaluate a possibility of a combination of our baseline with CMA-ES in Section \ref{sec:pretraining-retraining}. Finally, in Section \ref{sec:nonlinearity-inversion} we propose a technique to estimate the complex-valued states of the reservoir using only the final photodiode by running a predetermined input sequence several times, while appropriately setting the readout weights. The estimated complex states can then be used for classic training algorithms on a digital computer and the resulting weights can be programmed on the actual readout.

\section{Methodology}
\label{sec:methodology}
We evaluate the suitability of our proposed training approaches by comparing the achieved bit error rates of simulated reservoirs over a wide range of input signal bitrates. In more detail, we train the readout weights of simulated passive photonic reservoirs with integrated optical readout to perform the 3-bit header recognition task. Our simulation setup builds upon the setup in \cite{Katumba2017}: Katumba et al. inject an intensity-encoded bit signal into a simulated optical circuit corresponding to an integrated 4x4 passive photonic reservoir using the swirl \cite{Vandoorne2014} architecture as as illustrated in Figure \ref{fig:electrical_readout}. Further, they convert the complex-valued optical signals at each of the 16 nodes of the reservoir, from the optical into the electrical domain using a previously introduced photodetector model that takes into account the responsitivity, bandwidth and background noise of an integrated optical photodetector. Finally, they arrange the 16 sampled electrical signals into a reservoir state matrix and detect 1 bit delayed XOR bit patterns in the bit sequence which has previously been fed into the simulated photonic reservoir. In this work, contrary to Katumba et al., we do not convert all the reservoir's node signals into the electrical domain prior to training. Instead we assume an integrated optical readout operating directly on the optical signals of the reservoir capable of weighting and summing signals which are complex-valued by nature. By tuning the weights of this integrated optical readout, which is simulated on the functional level, we are able to perform meaningful computations by composing an optical signal from the signals occurring at the individual reservoir nodes. This optical signal is then converted from the optical into the electrical domain using the detector model used in \cite{Katumba2017}. 

In more detail, we use Caphe \cite{Fiers2012} to simulate an optical circuit corresponding to an integrated 4x4 passive photonic reservoir using the swirl \cite{Vandoorne2014} architecture. We subsample the intensity-modulated input signal 24 times per bit period, smoothen it using a single-pole low-pass filter, and simulate the response of our optical integrated swirl circuit using Caphe. We obtain a sampled complex output signal of each reservoir node as result, denoting amplitude and phase of the optical signal at that node at a certain instant in time. We arrange the sampled complex optical signals at each reservoir node as a complex state-node matrix $\matr{X} \in \mathcal{C}$ and simulate the integrated optical readout by computing an inner product between this matrix and a complex weight vector $\vec{w}$, which represents the complex optical weights. The resulting complex-valued signal is fed into a photodetector model to obtain the electrical output signal of the integrated photonic reservoir. This output power signal is then downsampled at a predetermined optimal sampling point and thresholded in order to obtain a clean binary output bit sequence. We place a threshold $T$ in the middle of the signal range of our output signal $y[n]$ as
\begin{equation}
T=\mathrm{P}_{5}(y[n]) + (\mathrm{P}_{95}(y[n]) - \mathrm{P}_{5}(y[n]))/2,
\end{equation}
where $\mathrm{P}_{5}(y[n])$ and  $\mathrm{P}_{95}(y[n])$ are the 5th and 95th percentile of $y[n]$ respectively.
We simulate the reservoir's optical readout as 
\begin{equation}
\label{eq:readout-model-methodology}
\vec{y} = \sigma(\matr{X}   \vec{w}),
\end{equation}
where $\matr{X} \in \mathcal{C}^{N \times F}$ is the matrix of complex reservoir states, containing $N$ samples of the complex signal occurring at $F$ reservoir nodes of an integrated passive photonic reservoir. $\vec{w} \in \mathcal{C}^{F \times 1}$ is a vector holding the complex weights of the integrated optical readout. $\sigma(\vec{a}): \mathcal{C}^{N} \rightarrow \mathcal{R}^{N}$ is the mapping from the optical to the electrical domain that is realized by the photodetector. In this paper, the photodetector model of \cite{Katumba2017} is used for all simulations and experiments.
This model computes the electric current of a sampled complex signal $\vec{a}$ as 
\begin{equation}
\vec{i}(\vec{a}) = R   |\vec{a}|^2,
\end{equation}
where $R$ is the responsitivity of the photodetector.
Thereafter a zero-mean Gaussian noise vector  $\mathbf{n}$  with a variance $\sigma_{n}^{2}$ is added to $\vec{i}(\vec{a})$.   
The variance $ \sigma_{n}^{2}$ is computed as 
\begin{equation}
\label{eqn:detector_noise}
\sigma_{n}^{2} = 2qB(\langle I \rangle + \langle I_{d} \rangle) + 4k_{B}TB/R_{L}  
\end{equation}
where $q$ is the elementary particle charge, $B$ is the bandwidth of the photodetector, $\langle I \rangle = \frac{1}{N} \sum_{n}^N i[n]$ is the photocurrent, $\langle I_{d} \rangle$ is the dark current, $k_{B}$ is the Boltzmann constant, $T$ is the temperature and $R_{L}$ is the load impedance of the photodetector. In our simulations we set $R=0.5 \frac{\mathrm{A}}{\mathrm{W}}$, $B=25\thinspace \mathrm{GHz}$, $\langle I_{d} \rangle =   0.1 \thinspace \mathrm{nA}$, $T=300 \thinspace  \mathrm{K}$ and $R_L=1 \thinspace  \mathrm{M\Omega}$. 
 Finally, to model the limited bandwidth $B$ of the integrated optical detector, a fourth-order Butterworth low-pass filter is applied to the resulting output signal.     

We fix the delay time between any two connected nodes in our simulated reservoir to $62.5$ ps and assume the waveguide loss to be a rather pessimistic 3 dB/cm. We inject the input signal into the reservoir through nodes 5, 6, 9 and 10, where node indices are ordered row by row and from left to right. This input node configuration offers a good trade-off between performance and wiring effort \cite{Katumba2017}.

As mentioned before, we use intensity modulation to encode our bit patterns into the optical signals sent to the simulated reservoir. In more detail, we distribute a total power $p_{total}=0.1$ Watt over the 4 input nodes, which leaves the maximal amplitude power per node input signal at $\frac{0.1 \mathrm{W}}{4} =0.025 \mathrm{W}$. Moreover, we have found that training our reservoirs with a bias term is beneficial to classifier performance. Thus we implement an additional optical bias line which does not lead into the reservoir, but directly into the integrated optical readout (see Figure \ref{fig:optical_readout}). This line carries a constant input power of $0.02 \mathrm{W}$. Every training algorithm in this paper treats the bias like an additional reservoir state signal, with the exception that the bias signal is not regularized during training. Assuming an equal occurrence probability of symbols 1 and 0 in the input signal, the average signal power to drive the reservoir amounts to $0.1 \mathrm{W} \cdot 0.5  + 0.02 \mathrm{W}=0.07 \mathrm{W}$. The power consumption for actual implementations of an integrated optical readout with possibly active weighting elements needs of course also to be taken into account, but is beyond the scope of this work.

Since we seek to find a training algorithm that works well over a wide range of dynamics, we assess the performance of our classifiers by training integrated photonic reservoirs excited by input signals over a wide range of bitrates. We sweep the bit rate of the input signal in $1$ Gbps steps between $1$ and $31$ Gbps. 

As a machine learning task to assess the performance of our classifiers, we use the header recognition task. We expect the reservoir to present $1$ at the output whenever a certain sought header bit sequence occurs in the input signal and $0$ otherwise.

More precisely, given an input signal $u[n]$ and a predefined header bit pattern $h[n]$ we define our ideal desired signal $d_{\text{ideal}}[n]$ to be
\begin{equation}
d_{\text{ideal}}[n] = 
   \begin{cases}
      1 & \text{if} \sum_{m=0}^{M-1} \llbracket h[(M-1)-m] = u[n-m]\rrbracket = M  \\
      0 & \text{else }
    \end{cases} \\
\end{equation}
where the notation above is Iversons bracket notation, defined as
\begin{equation}
\llbracket P \rrbracket = 
   \begin{cases}
      1 & \text{if } P \quad \text{true}  \\
      0 & \text{else }
    \end{cases}
\end{equation}
and M is the length of $h[n]$ in bits. We start out setting $M=3$ and $h[n]=\delta[n] + \delta[(n-2)]$, where $\delta$ denotes the dirac delta function, thus seeking to detect the header bit pattern "101". We use this pattern throughout this work as we discuss our findings. Thereafter, to demonstrate that our introduced approaches exhibit consistent performance on the task, we train classifiers to detect all possible 3 bit headers. While it is more common in recent literature  \cite{qin2017numerical, Vandoorne2014} to compute errors in a winner-takes-all fashion jointly for all trained classifiers and patterns, we analyze the detection performance of each header bit pattern individually. We argue that individual classifiers can easier be analyzed employing the latter approach, especially when investigating reservoir performance over a large range of bitrates.

We train our reservoir readouts using a modified version of the ideal desired signal
\begin{equation}
d[n]=d_{\text{ideal}}[n] \cdot   p_{\text{total}},
\end{equation}
 where again $p_{\text{total}}=0.1 \thinspace \mathrm{W}$, the maximal attainable power of the input signal.

We generate 10010 random bits as training data as well as 10010 random bits of test data. Since we remove the reservoir signals generated by the first 10 input bits for both training and test states in order to minimize the impact of transients, this leaves us with 10000 bits of training data as well as 10000 bits of test data. To account for manufacturing variations, in general, we simulate each reservoir 10 times with identical train and test input, but different random phase configurations of the waveguides between nodes, as well as of the waveguides feeding the input signals to the nodes. Subsequently we train the output weights of the simulated readout for each instance. 

To compare the performance of our trained classifiers, we use the bit error rate (BER). The BER is defined as
\begin{equation}
e_{\text{bit}}=\frac{1}{N} \sum_{n=1}^N \llbracket y_{T}[n] \neq d_{\text{ideal}}[n] \rrbracket,
\end{equation}
where $y_{T}[n]$ is the subsampled, thresholded output signal of the reservoir and $d_{\text{ideal}}[n]$ is again the ideal desired signal.
With 10000 bits of test data, the minimal detectable bit error rate with a confidence level of $\approx 90\%$ is $10^{-3}$ \cite{Jeruchim1984Techniques}. 
Before computing the bit error rate on the test data, for all trained classifiers we determine the optimal sampling point minimizing the bit error rate on the train data, which we then use to downsample the test prediction and compute the bit error rate as described above. We have found that this additional step leads to more robust results for all classifiers investigated.

\section{Baseline: Complex-Valued Ridge Regression}
\label{sec:complex-ridge}

Consider again the model of an integrated optical readout as introduced in Section \ref{sec:methodology}, defined as 
\begin{equation}
\label{eq:readout-model}
\vec{y} = \sigma(\matr{X}   \vec{w}).
\end{equation}
$\matr{X} \in \mathcal{C}^{N \times F}$ is the matrix of complex reservoir states, containing $N$ samples of the complex signal occurring at $F$ reservoir nodes of an integrated passive photonic reservoir. $\vec{w} \in \mathcal{C}^{F \times 1}$ is a vector holding the complex weights of the integrated optical readout. $\sigma(\vec{a}): \mathcal{C}^{N} \rightarrow \mathcal{R}^{N}$ is the mapping realized by the photodetector of the readout.

Contrary to classic reservoir readouts, the readout weights of this model do not lie after the reservoir's nonlinear detector function, but before it. This implies that any result of the dot product $\matr{X}   \vec{w}$ of the model will be passed through this nonlinear function. Consequently, if we train our readout weights in a classical way, using ridge regression \cite{hoerl1970ridge}, the desired product vector $\vec{d}$ of states and readout weights will be transformed by the detector output function. In more detail if we train $\vec{w}$ as  
\begin{equation}
\label{eq:complex-ridge-vanilla}
\vec{w} = (\matr{X}^H \matr{X} + \alpha^2 I)^{-1} \matr{X}^H \vec{d},
\end{equation}
where $\alpha \in \mathcal{R}$ is the regularization strength and $\matr{I} \in \mathcal{R}^{F \times F}$ is the identity matrix. Assuming $\vec{w}$ is ideal,
\begin{equation}
\sigma(\matr{X}   \vec{w}) = \sigma(\vec{d}).
\end{equation}
Therefore, at the model output, we obtain
\begin{equation} 
\vec{y} = \sigma(\matr{X}   \vec{w}) = \sigma(\vec{d}).
\end{equation}

Since we would like our model to output $\vec{d}$ rather than $\sigma(\vec{d})$, we need to find an approximate inversion of the readout nonlinearity $\hat{\sigma}^{-1}$ to invert $\sigma$ such that 
\begin{equation}
\sigma(\hat{\sigma}^{-1}(\vec{d})) \approx \vec{d}.
\end{equation}  
We therefore train $\vec{w}$ as 
\begin{equation}
\label{eq:complex-ridge}
\vec{w} = (\matr{X}^H \matr{X} + \alpha^2 I)^{-1} \matr{X}^H \hat{\sigma}^{-1}(\vec{d}),
\end{equation}

We can approximate the detector model in \cite{Katumba2017} (for details, see Section \ref{sec:methodology}) as a function in closed form, if we neglect its bandlimiting lowpass filter for simplicity :
\begin{equation}
\sigma(\vec{a}) = R   |\vec{a}|^2 + \vec{n}.
\end{equation} 

Again, $R$ denotes the responsitivity of the photodetector and $\vec{n}$ is a noise vector. One can see that $\sigma(\vec{a})$ cannot be inverted exactly, primarily since we take the absolute value of $\vec{a}$, and also due to the added, unknown noise vector. Nevertheless we can approximate the inverse of the detector function above as
\begin{equation}
\label{eq:abs-estimation}
\hat{\sigma}^{-1}(a) = \sqrt{\frac{a}{R} \mathstrut}.
\end{equation}

By doing so, we minimize the sum of squared errors
\begin{equation}
\label{eq:sum-of-sq-errors}
\sum_{n=0}^{N} \left[(\vec{x^{(n)}}   \vec{w}) - \sqrt{\frac{d[n]}{R}}\right]^2
\end{equation}
where we denote the row vector with index $n$ of $\matr{X}$ as $\vec{x^{(n)}}$.

While this approach obviously can not be used on actual devices due to the fact that the optical signals on the chip are not observable, its similarity to classic, real-valued training approaches makes it a suitable candidate to be used as a baseline to assess and compare novel training approaches for actual integrated reservoirs. Whenever we refer to this baseline we call it plainly \emph{complex-valued ridge regression}. Whenever we train classifiers using complex-valued ridge regression, we use 5-fold cross validation to find a suitable regularization parameter $\alpha$ for an optimal training result.

\begin{figure}[!t]
\centering
\includegraphics[width=3.2in]{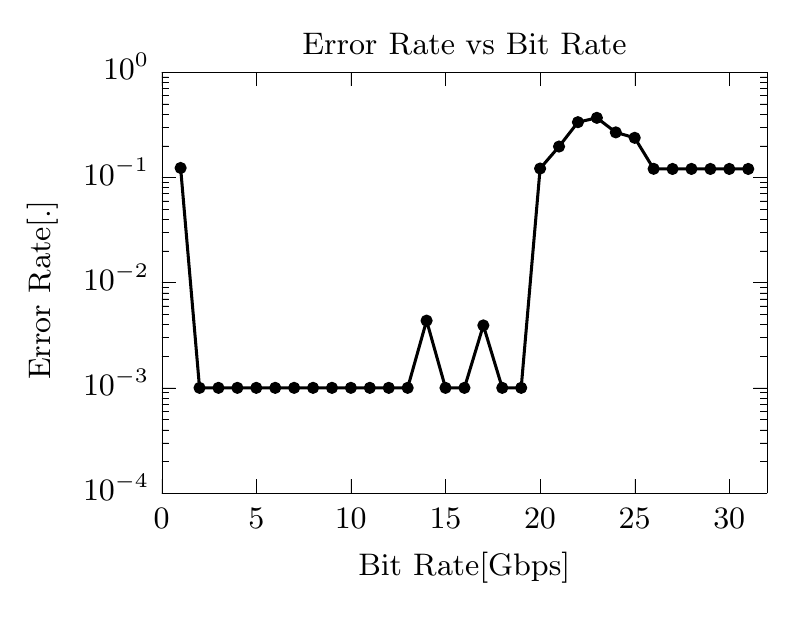}

\caption{Bit error rates as a function of input signal bitrate for our baseline approach, complex-valued ridge regression. Classifiers have been trained to perform 3 bit header recognition (pattern 101) on a 4x4 passive photonic swirl. All results are averaged over 10 different reservoirs, the minimal detectable error rate is $10^{-3}$.}
\label{fig:baseline}
\end{figure}

We assess the performance of this baseline by training integrated photonic reservoirs for different input signal bitrates on the 3-bit header recognition task. Figure \ref{fig:baseline} shows the achieved bit error rates.


As one can see, our proposed baseline works well for bitrates between 2 and 20 Gbps with minor increases of the error for bitrates at 14 Gbps and 17 Gbps.

\section{Black-Box optimization}
\label{sec:black-box-optimization}
Since the states of our integrated photonic reservoir are not observable, a straight-forward approach is to train the readouts using a black-box optimization approach. Among these, Convergence Matrix Adaption - Evolution Strategy (CMA-ES) \cite{hansen2001completely} appears to be a suitable candidate, since it usually deals well with non-convex search spaces, which are likely to occur for our problem since we perform optimization in the complex domain. Evolution strategies solve optimization problems in an iterative way using a repeated two-step process based on of the variation and selection of possible solutions. In the variation step, new candidate solutions are generated by adding noise to the currently known best solution. This noise is drawn from a normal distribution called the mutation distribution. Thereafter, newly generated candidate solutions are evaluated using a predefined cost function. Candidate solutions which perform better than the previously best known solution are selected and used to update the parameters of the mutation distribution. Using the updated best known solution as well as the updated mutation distribution, the steps described above are repeated. The overall process can be repeated until a satisfactory solution to the given optimization problem has been found. The exact rules how the mutation distribution is updated vary inbetween evolution strategies. CMA-ES adapts the mean value and covariance matrix of the used mutation distribution such that the probability to draw updates which lead to previously selected solutions increases. The underlying rationale is that conducting updates which lead to good solutions in the past might result in even better solutions when performed again. For details see \cite{hansen2001completely}.  

 Integrated photonic reservoirs can be trained using CMA-ES by transferring a candidate solution suggested by the algorithm  to the reservoir's readout weights, presenting the training bit sequence to the reservoir, computing a chosen loss function from the reservoir's subsequent output and feeding that error measure back to the CMA-ES algorithm. The algorithm traverses the loss function space suggesting new candidates to minimize the loss function. We encode the real and imaginary parts of the complex weight vector $\vec{w} \in \mathcal{C}$ into a real-valued vector 
\begin{equation}
\vec{w}'=\begin{pmatrix} \text{Re}(\vec{w})  \\ \text{Im}(\vec{w}) \end{pmatrix}
\end{equation}
 prior to handing it to the CMA-ES algorithm. The inverse transformation is applied to weight vectors suggested by the algorithm before setting them to the readout. We initialize the algorithm with a zero vector $w'_{0}=\vec{0}$ and perform sweeps over the initial variance in steps of one decade between $10^{-5}$ and $10^2$. As recommended in  \cite{hansen2016cma}, we set the population size to  $4+ \lfloor 3 \log(F) \rfloor$. Note that we might achieve better results by cross-validating initial variance and population size, from which we refrain due to the typical very long training times of CMA-ES. As an objective function, we minimize the sum of squared errors
\begin{equation}
e_{SSE}= \sum_{n=1}^{N} (y[n] - d[n])^2.
\end{equation}

\begin{figure}[!t]
\centering
\includegraphics[width=3.2in]{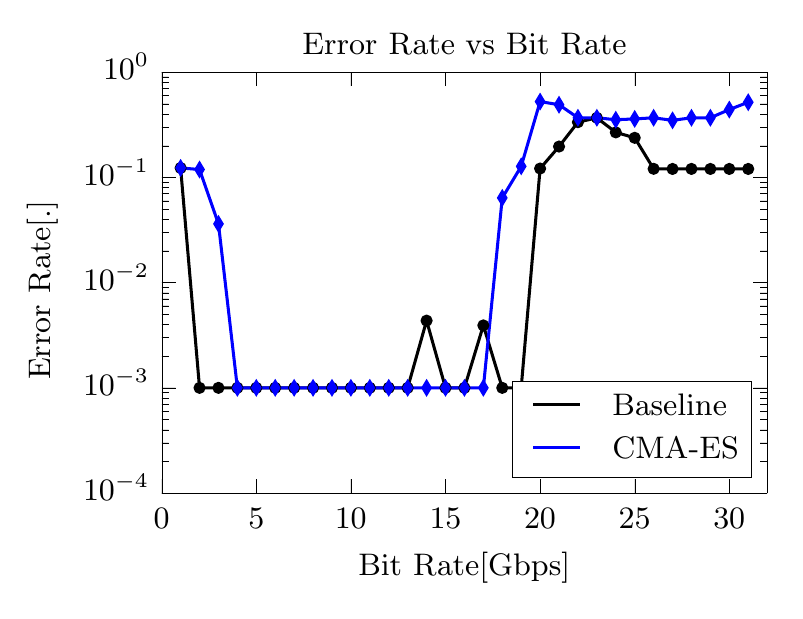}

\caption{Bit error rates as a function of input signal bitrate for the black-box approach, CMA-ES. Baseline approach is shown for comparison. Classifiers have been trained to perform 3 bit header recognition (pattern 101) on a 4x4 passive photonic swirl. All results are averaged over 10 different reservoirs, the minimal detectable error rate is $10^{-3}$.}
\label{fig:cmaes}
\end{figure}

We assess the performance of CMA-ES by training integrated photonic reservoirs with different input signal bitrates on the 3-bit header recognition task. Figure \ref{fig:cmaes} shows the resulting error rates (results are averaged over 10 different reservoirs).

\begin{figure}[!t]
\centering
\includegraphics[width=3.2in]{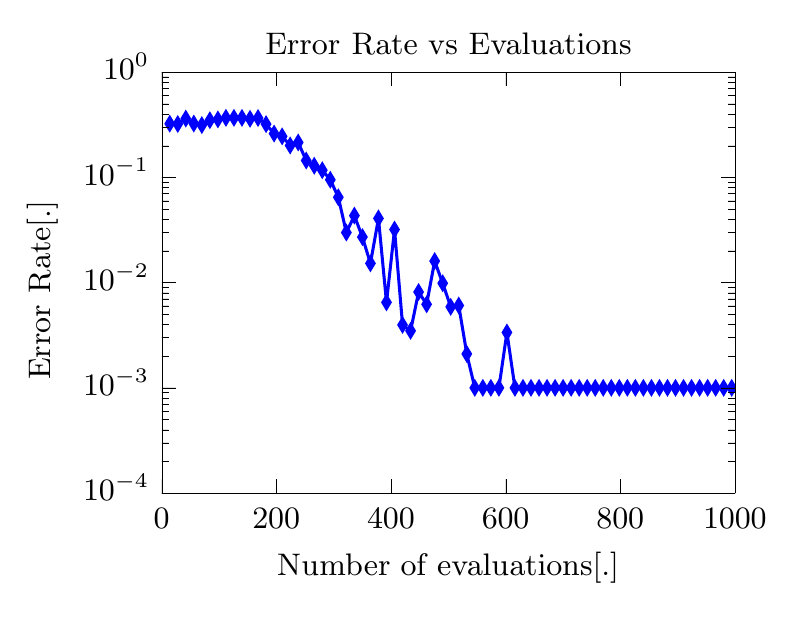}

\caption{Bit error rate as a function of iterations over the training data for CMA-ES. Classifiers have been trained to perform 3 bit header recognition (pattern 101) on a 4x4 passive photonic swirl. All results are averaged over 10 different reservoirs, the minimal detectable error rate is $10^{-3}$.}
\label{fig:experiment-convergence}
\end{figure}

As one can see, CMA-ES performs only slightly worse than the baseline, achieving the minimal detectable bit error rate of $10^{-3}$ for bitrates between 4 and 17 Gbps. Therefore, it is {\em in principle} capable of training integrated photonic reservoirs. However, because CMA-ES typically involves high training time and requires many iterations of the input data, we run an additional experiment investigating its convergence behavior. We simulate a passive photonic reservoir and train it with CMA-ES where we record the error rate at each iteration. We again average our results over 10 simulated reservoirs driven by a bit rate of 10 Gbps to obtain the graph seen in Figure \ref{fig:experiment-convergence}. 

The results of this experiment show that the full input training sequence needs to be presented about 500 times before CMA-ES reaches satisfactory results. Since a short training process on the actual hardware is mandatory for our devices in order to obtain mass-market maturity, a way to drastically reduce the number of necessary iterations over the input is necessary. Since stand-alone CMA-ES training converges too slowly, a promising alternative is the pretraining of models in simulation and the refining them using CMA-ES on the actual devices to speed up the training process. We investigate the feasibility of such a pretraining-retraining approach in the next section.

\section{Pretraining in simulation}
\label{sec:pretraining-retraining}

As mentioned in Section \ref{sec:introduction}, training reservoirs in simulation allows full observability. However the resulting weights are not directly transferable to hardware, due to the fabrication tolerances of integrated photonic reservoirs. However, if the changes in reservoir output signals due to process variations are small enough, a pretraining approach could be useful if weights trained in simulation still perform considerably better than random weights. In such a case, one could use them to initialise, e.g., the training with CMA-ES, leading to much faster training convergence. In this section, we quantify how phase variability of in our reservoirs affects the quality of weights trained in simulation. Our approach is as follows:

\begin{enumerate}
\item We train the weights of a {\em nominal} simulated reservoir with integrated readout using complex-valued ridge regression. We choose the bit rate of the input signal to be $5$ Gbps, based on Figure \ref{fig:baseline}. 
\item We create simulation models of $10$ possible physical instances of the previous reservoir that reflect possible waveguide phase variations in a physical reservoir due to  manufacturing. This is done by adding random variations $ \mathbf{\eta}_i $ and $\mathbf{\eta}_c$ to the phase configurations of input and connection waveguides of the nominal reservoir. The phase variations are drawn from a uniform distribution $\mathcal{U}(0,b)$, where $b$ represents the maximal \emph{phase perturbation} in each waveguide. 
\item We reapply the previously trained readout weight vector to each of these physical reservoir models and record the resulting bit error rate. 
\end{enumerate}

This procedure is repeated for for all perturbations $b \in \{0.1 \pi,0.2 \pi,0.3 \pi,0.4 \pi,0.5 \pi,0.6 \pi,0.7 \pi,0.8 \pi,0.9 \pi,\pi \}$. In addition, for each value of $b$, the resulting bit error rates are averaged across $10$ different instances of the nominal reservoir (each with different random phases of all connections).  
\begin{table}[!t]
\centering
\begin{tabular}{|c| c |c |c| c |}
 \hline
\multicolumn{5}{|c|}{Maximum Phase Perturbation} \\
 \hline
$0$ & $0.1 \pi$ & $0.2 \pi$ & $ 0.3 \pi$ & $ > 0.3 \pi$ \\   
 \hline
$ < 0.001 $ & $0.506$ & $0.643$ & $0.673$ & $\approx 0.70$    \\
 \hline
\end{tabular}
\caption{ Bit error rate for the 3 bit header recognition task (pattern 101) for increasing phase perturbations in the reservoir's waveguides (readout trained in simulation with complex-valued ridge regression).}
\label{fig:table-pt-in-simulation}
\end{table}

Table \ref{fig:table-pt-in-simulation} summarises the results. The bit error rate already increases by two orders of magnitude for amounts of random phase noise that are currently well below fabrication tolerances. This renders a pretraining-retraining approach challenging. In the next section, we therefore pursue a radically different approach, in which we estimate the reservoir's states through the available photodetector in order to train a weight vector from the actual states on a digital computer. This weight vector can then be transferred back to the hardware where no significant increase in error is to be expected.

\section{Nonlinearity Inversion}
\label{sec:nonlinearity-inversion}
Consider again the model of an integrated optical readout as introduced in Section \ref{sec:methodology}, defined as 
\begin{equation}
\label{eq:readout-model-nlinv}
\vec{y} = \sigma(\matr{X}   \vec{w}),
\end{equation}
the closed form approximation of the photodetector function
\begin{equation}
\sigma(\vec{a}) = R   |\vec{a}|^2 + \vec{n},
\end{equation} 
 as well as its approximate inversion
\begin{equation}
\label{eq:abs-estimation-nlinv}
\hat{\sigma}^{-1}(a) = \sqrt{\frac{a}{R} \mathstrut},
\end{equation}
introduced in Section \ref{sec:complex-ridge}.

$\hat{\sigma}^{-1}$ can be used to estimate the reservoir state matrix $|X|$ when it cannot be observed directly.

Indeed, taking a closer look at Equation \ref{eq:readout-model-nlinv}, one can see that it is possible to observe the powers of the state matrix $\matr{X}$ through an appropriate selection of the input weight vector $\vec{w}$. If we choose 
\begin{equation}
w=\begin{pmatrix} 1 \\ 0 \\ \vdots \\ 0 \end{pmatrix}
\end{equation}
as a weight vector and present an input signal $u[n]$ to the input of the integrated photonic reservoir, we obtain
\begin{equation}
\vec{y} =\sigma(\vec{x_1}) =R   |\vec{x_1}|^2 + \vec{n},
\end{equation}
at the output of the integrated readout, where $\vec{x_1}$ is the first column of the state matrix $\matr{X}$.
Since we can observe $ R    |\vec{x_1}|^2 + \vec{n}$, we can estimate the modulus $|\vec{x_1}|$ of
\begin{equation}
\vec{x_1}= |\vec{x_1}|   \exp(\mathrm{j} \arg(\vec{x_1})).
\end{equation}

If we neglect $\vec{n}$, assuming we know $R$, we can apply $\hat{\sigma}^{-1}$ to $\vec{y}$
and estimate $|\vec{x_1}|$ as

\begin{equation}
\hat{|\vec{x_1}|}=\hat{\sigma}^{-1}(R   |\vec{x_1}|^2 + \vec{n}) = \sqrt{ |\vec{x_1}|^2 + \frac{\vec{n}}{R}}.
\end{equation}

Since the above operation is an approximate inversion of the nonlinearity of the photodetector, we call our method \emph{nonlinearity inversion}.

\begin{figure}[!t]
\centering
\includegraphics[width=3.2in]{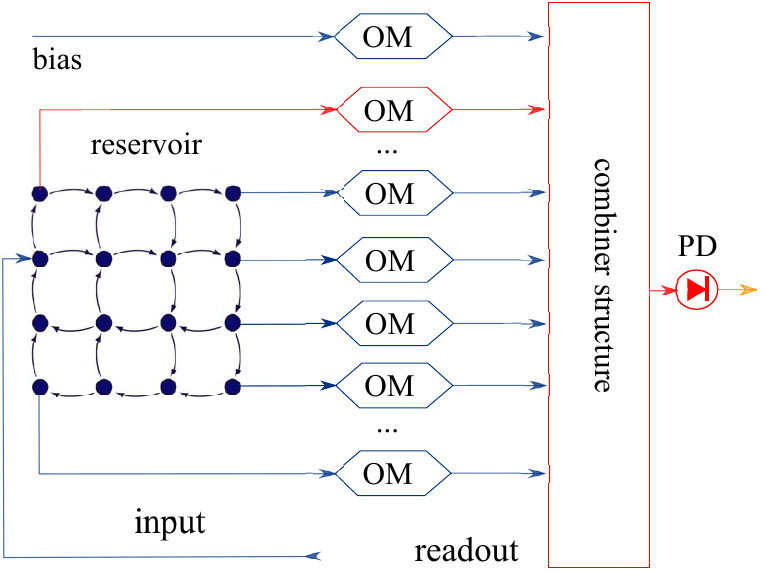}
\caption{Illustration of modulus (light intensity) observation procedure. The weight highlighted in red is set to $1$, all remaining weights are set to $0$, the observable output is the square of the modulus of the corresponding reservoir state.}
\label{fig:config_meas_amps}
\end{figure}
Repeating this procedure for every channel of the readout/column of the state matrix, we are able to estimate the moduli of state values in the state matrix $|X|$. See Figure \ref{fig:config_meas_amps} for illustration.

Nevertheless, in order to tune the weights of the integrated optical readout, we need information about the arguments (the phases) of $|X|$ as well. Absolute phase information about the states is lost as soon as it passes the photodetector. However, only the relative phases between states are of interest since they influence the output sum of the readout.

Consider two given complex state values $x_{(t,k)}$ and $x_{(t,l)}$ at a certain instant in time $t$, with moduli $P_k$ and $P_l$ as well as the modulus $P_{kl}$ of their sum $P_{kl}   \exp(\mathrm{j} \phi_{kl}) = P_k   \exp(\mathrm{j} \phi_k) + P_l   \exp(\mathrm{j} \phi_l)$. The absolute value of their relative phase difference $\phi_{kl}=\phi_k-\phi_l$ can be computed using the phase estimation equation (see Appendix \ref{sec:phase-est-deriv} for derivation) 
 \begin{equation}
\label{eq:phase-estimation}
 |\phi_{kl}| = \arccos\left(\frac{P_{kl}^2 - (P_k^2+P_l^2)}{2   P_k  P_l}\right).
\end{equation}

Due to the fact that $\arccos(x)$ is injective in the sense that for any given input $x$ there are two possible solutions on the interval $[-\pi, \pi]$, using Equation \ref{eq:phase-estimation} we can only find the absolute value $|\phi_{kl}|$, while the sign of $(\phi_{kl})$ remains unknown.

To resolve this issue, one performs this estimation of $\phi_{kl}$ twice with different values $P_{kl}$ and $P_{kl}'$: While $P_{kl}$ is being computed as shown before, a phase difference of $\frac{\pi}{2}$ is added to state $k$ such that 
\begin{equation}
\begin{split}
P_{kl}'  =& |P_k   \exp(\mathrm{j} \phi_k)   \exp(\mathrm{j}  \frac{\pi}{2}) + P_l   \exp(\mathrm{j}  \phi_l)|^2
 \\=& |P_k   \exp(\mathrm{j}  \phi_k') + P_l   \exp(\mathrm{j}  \phi_l)|^2. 
\end{split}
\end{equation} 

By comparing the estimates of $|\phi_{kl}|$ and
 \begin{equation}
 |\phi_{kl}'| = \arccos\left(\frac{P_{kl}'^2 - (P_k^2+P_l^2)}{2   P_k  P_l}\right),
\end{equation}
we are able to infer the sign of $\phi_{kl}$.
\begin{equation}
\label{eq:phase-sgn}
\phi_{kl}=
   \begin{cases}
       |\phi_{kl}| & \text{if }  |\phi_{kl}'| \in [0, \frac{\pi}{2}]  \\
      -|\phi_{kl}| & \text{else }
    \end{cases} \\
\end{equation}

\begin{figure}[!t]
\centering
\includegraphics[width=3.2in]{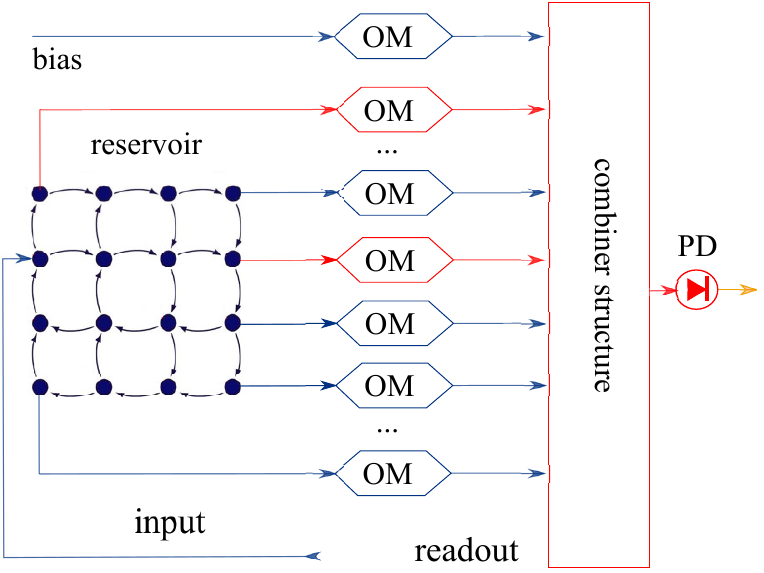}
\caption{Illustration of phase estimation procedure. The weights highlighted in red are set to $1$, the remaining weights are set to $0$. This results in the power of the summed states at the output, which allows us to calculate the argument (phase angle) between states of the highlighted channels via the relationship of the sum power and the powers of the individual states obtained in the previous step.}
\label{fig:config_meas-phases}
\end{figure}

(see Appendix \ref{sec:phase-est-deriv} for proof). For illustration, see Figure \ref{fig:config_meas-phases}.

Since
\begin{equation}
|P_k   \exp(\mathrm{j}  \phi_k) + P_l   \exp(\mathrm{j}  \phi_l) | = |P_k + P_l   \exp(\mathrm{j}  \phi_{kl}) |
\end{equation}
and thus in consequence
\begin{equation}
\begin{split}
|P_k   \exp(\mathrm{j}  \phi_k) + \sum^{\forall q \neq k} P_q   \exp(\mathrm{j}  \phi_q) | = 
\\|P_k + \sum^{\forall q \neq k} P_q   \exp(\mathrm{j}  \phi_{kq}) |, 
\end{split}
\end{equation}
we can pick a certain state $x_{(t,k)}$ as reference state and determine the relative phase $\phi_{kq}$ between the value of this state $x_{(t,k)}$  and the value of every other state $x_{(t,q)}$.

After having estimated each reservoir state this way, we apply complex-valued ridge regression to find the optimal weights for our reservoir.

In summary, after presenting the same input $3F-2$ times, with $F$ being the number of output channels, we are able to measure the full complex time evolution of each of the $F$ output channels, even though we only have a single detector. This information can then be used to calculate the required weights in a single pass in software on a digital computer, after which the computed weights are transferred back to the readout. For our simulated 16 node reservoirs, taking into account the bias channel, which can be treated like any other readout channel, this requires to present the input data $3F-2=3 \times (16+1)-2=49$ times to the reservoir. As we have seen in Section \ref{sec:black-box-optimization}, $3F-2$ measurements is much less than what is typically necessary using CMA-ES. In addition, it is also a deterministic number, in contrast to a black-box optimisation technique, for which it is hard to determine beforehand how many iterations will be needed.

\begin{figure}[!t]
\centering
\includegraphics[width=3.2in]{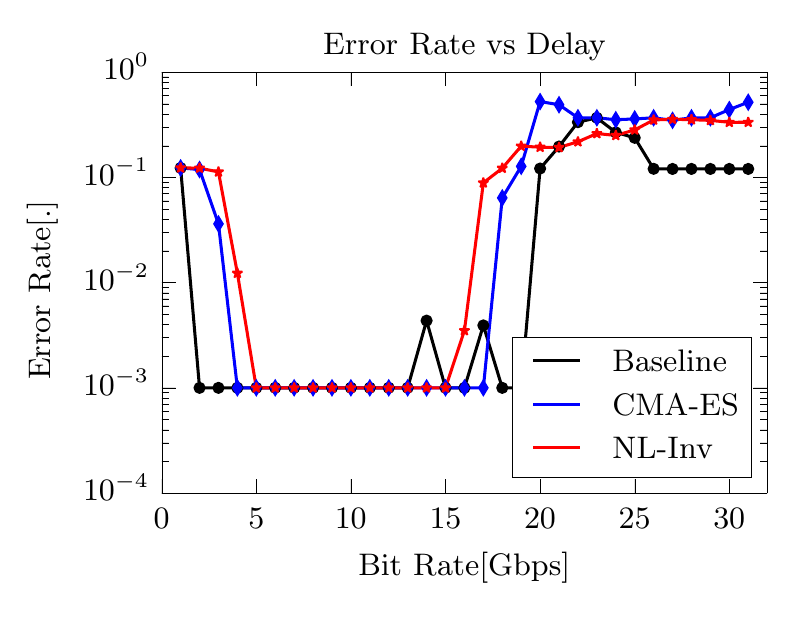}
\caption{Bit error rates as a function of input signal bitrate for our proposed approach, nonlinearity inversion. Baseline and CMA-ES  approaches are shown for comparison. Classifiers have been trained to perform 3 bit header recognition (pattern 101) on a 4x4 passive photonic swirl. All results are averaged over 10 different reservoirs, the minimal detectable error rate is $10^-{3}$.}
\label{fig:nlinv}
\end{figure}

In our experiments, we simulate the repeated measurements for the nonlinearity inversion procedure by setting the corresponding rows of the weight vector in the readout model according to our estimation procedure. For every setting of the weight vector we apply the complete model of our readout as described in Section \ref{sec:methodology}. We collect the corresponding output signal from the detector and replace any samples of the output signal smaller than $0$ (which might occur due to noise or ringing of the bandlimiting filter of the photodetector) with $0$. Thereafter $\matr{X}$ is estimated using Equations \ref{eq:abs-estimation-nlinv}, \ref{eq:phase-estimation} and \ref{eq:phase-sgn}. 
We again train integrated photonic reservoirs with input bit rates between nodes on the 3-bit header recognition task to asses the performance of our proposed method. Figure \ref{fig:nlinv} shows the bit error rate as function of the input signal bit rate.

\begin{figure}[!t]
\centering
\includegraphics[width=3.5in]{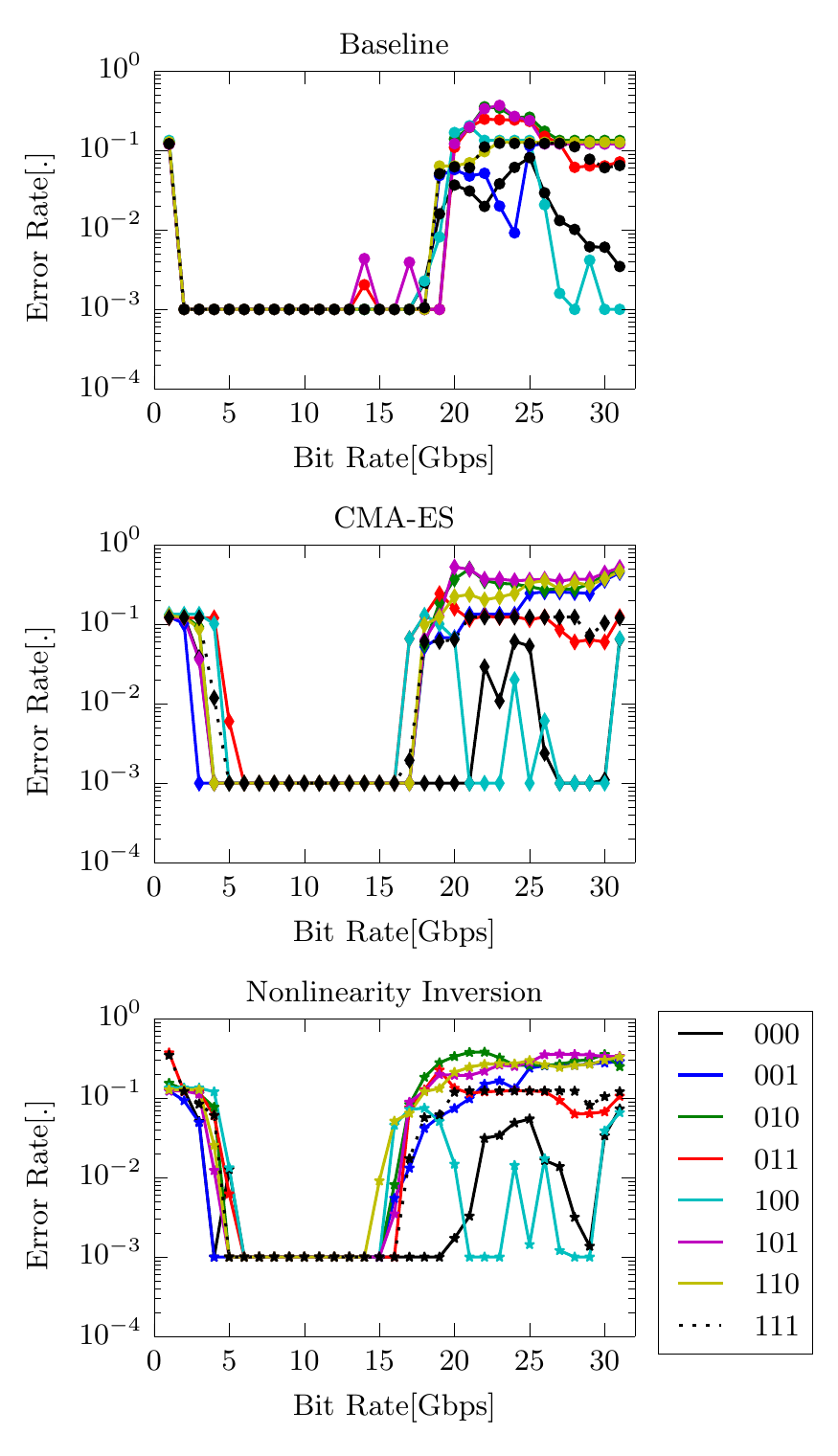}
\caption{Bit error rates as a function of input signal bitrate for all possible 3 bit headers as performed by our baseline, CMA-ES and nonlinearity inversion approaches. Classifiers have been trained to perform 3 bit header recognition on a 4x4 passive photonic swirl. All results are averaged over 10 different reservoirs, the minimal detectable error rate is $10^-{3}$.}
\label{fig:all_patterns}
\end{figure}

The nonlinearity inversion approach performs only slightly worse than the the CMA-ES approach and the complex-valued ridge regression baseline. It is remarkable though, that for some bitrates, for instance at 14 Gbps, the nonlinearity inversion approach outperforms the baseline. As the nonlinearity inversion approach operates on an estimate of the states used by the baseline, one would expect it to perform at best just as well as the baseline. A possible explanation for this phenomenon is that the noise introduced by the detector model in the estimation step acts as an additional regularizer for training. To provide a comprehensive picture on the investigated training approaches, we have performed the last experiment measuring the performance when recognizing all possible bit patterns in a 3 bit header. Figure \ref{fig:all_patterns} shows the results for all approaches discussed in this work. The results for all headers are largely consistent with the results we already observed for bit pattern '101':
Our baseline approach performs consistently well over a wide range of bitrates for all bitpatterns, approximately between 2 and 18 Gbps, with minor increases in error at 14 and 17 Gbps. For CMA-ES on the other hand, the performance varies more in between bit patterns, where it performs very well for some patterns such as '000' and '100 'and slightly less well for others such as '011'. Nevertheless, CMA-ES attains the minimal error rate for all bitpatterns in a range between 6 and 16 Gbps. Interestingly, it manages to minimize the error rate for the '100' pattern between 20 and 22 Gbps where our baseline performs two orders of magnitude worse. This is consistent with the performance of our nonlinearity inversion approach for that pattern, which also manages to minimize the error rate in the same region contrary to the baseline. The overall performance across bitrates is slightly smaller than for the two former approaches, still, nonlinearity inversion minimizes the error rate for the given task on all possible header bit patterns on a range between 6 and 14 Gbps, and is thus almost on par with CMA-ES here. In a nutshell, since nonlinearity inversion shows only slightly worse performance than CMA-ES, while requiring significantly less ($3F-2=3 \times (16+1)-2=49$ times) iterations of the input data, it appears to be the most suitable among the training approaches for integrated photonic reservoirs that have been investigated in this paper. 

\section{Conclusion}
In this paper, we have assessed the suitability of different training methods for passive photonic reservoirs with integrated optical readout. The most important boundary condition for this type of hardware is the fact that the internal reservoir states are not directly observable in the electronic domain. In addition, the variability between individual devices makes it necessary to perform the training on the actual hardware. For this reason, our analysis has focused on the total training time while staying below the maximally allowed bit-error rate of $10^{-3}$. 

In the process, we found that CMA-ES delivers acceptable performance, but has to be ruled out as a practical training method due to its long training times. Due to the high fabrication tolerances of integrated photonic reservoirs, the use of pretraining-retraining as a way to reduce these training times does also not seem very promising. As an alternative, we have proposed a novel method for training such reservoirs which we call nonlinearity inversion. Our method essentially resolves the issue of limited observability of the states of an integrated passive photonic reservoir by estimating the reservoir's states through a single photodetector at its output. In more detail, we iterate over the training data several times while setting the weights according to a certain pattern. The recorded output signals allow us to estimate the amplitude and phase of the reservoir's states within $3F-2$ iterations over the training data. We have shown that this method performs as well as a classic training approach, which requires full observability, although for a more narrow bitrange. While the CMA-ES black box algorithm performs slightly better than our own method with respect to task performance, our method requires significantly fewer iterations over the input data. We conclude that nonlinearity inversion is a suitable candidate for training integrated passive photonic reservoirs.

While all our approaches are roughly comparable with respect to task performance, we find that the baseline approach still outperforms the remaining approaches. On has to take into account here though, that the baseline approach requires full observability and thus can not be implemented without considerable hardware effort which affects the scalability of our systems. In contrast, CMA-ES and our nonlinearity inversion approach exhibit slightly worse task performance, but are still applicable for a large range of bit rates. It is remarkable that our approach exhibits better results than the baseline at certain bit rates, especially since it operates on an estimate of the states used by the baseline. A possible explanation for this phenomenon is that the noise introduced by the detector model in the estimation step acts as an additional regularizer for training. In future work, we will further investigate the reasons behind this better performance in order to come up with a more accurate but still efficient way of training the weights based on the estimated states obtained from nonlinearity inversion. Moreover we will investigate the performance on actual hardware implementations of integrated optical readouts as soon as they are available. While we have assumed in this work that amplitude and phase of the readout are largely independently tunable, we are aware that this will not be perfectly realizable for all hardware implementations, especially considering nonvolatile tunable weights. We will address this problem in future work which could be met for instance by refining weights trained by nonlinearity inversion using reinforcement learning.

\appendices
\section{Derivation of the Phase Estimation Process}
\label{sec:phase-est-deriv}

Two given optical signals of our reservoir states $x_{(t,k)}$ and $x_{(t,l)}$ can be represented as

\begin{equation}
x_{(t,k)}=P_k  \exp(\mathrm{j} \phi_k)
\end{equation}
 and 
\begin{equation}
x_{(t,l)}=P_l  \exp(\mathrm{j} \phi_l)
\end{equation}
respectively. Given $P_k$, $P_l$ , as well as the squared sum of both signals
\begin{equation}
P_{kl}^2=|P_k \exp({\mathrm{j} \phi_k}) + P_l  \exp(j \phi_l))|^2
\end{equation}

we want to gain information on the state phases $\phi_k$  and $\phi_l$.
Applying Euler’s identity on the previous equation, separating real and imaginary parts and computing the absolute value gives us
\begin{equation}
P_{kl}^2 = [P_k \cos(\phi_k) + P_l \cos(\phi_l)]^2 + [P_k  \sin(\phi_k) + P_l \sin(\phi_l)]^2.
\end{equation}
After expanding the squared brackets, we get
\begin{equation}
\begin{split}
P_{kl}^2 =& P_k^2  \cos(\phi_k)^2+ 2 P_k P_l  \cos(\phi_l) \cos(\phi_k) +\\
& P_l^2  \cos(\phi_l)^2 + P_k^2  \sin(\phi_k)^2+ \\
& 2 P_k P_l \sin(\phi_l) \sin(\phi_k) +  P_l^2  \sin(\phi_l)^2. 
\end{split}
\end{equation}
By setting
\begin{equation}
\label{eq:square-id}
  \sin(\phi_l)^2 = 1 - \cos(\phi_l)^2
\end{equation}
and applying the trigonometric identities
\begin{equation}
\cos(\phi_l)   \cos(\phi_k) = \frac{1}{2}  \cos(\phi_l - \phi_k) +  \frac{1}{2} \cos(\phi_l + \phi_k) 
\end{equation}
and 
\begin{equation}
\sin(\phi_l)  \sin(\phi_k) = \frac{1}{2} \cos(\phi_l - \phi_k) -  \frac{1}{2} \cos(\phi_l + \phi_k), 
\end{equation}
we obtain
\begin{equation}
\label{eq:state_output_relation_final}
P_{kl}^2 = P_k^2 + P_l^2 + 2  P_k P_l \cos(\phi_l - \phi_k).  
\end{equation}
As expected, we can see in the last equation that the output of the system only depends on the difference of the input signal phases, but not the absolute phases themselves. Thus if we consider
\begin{equation}
|P_k + P_l \exp(\mathrm{j}(\phi_l - \phi_k))|^2
\end{equation}
we get
\begin{equation}
\begin{split}
& P_{k}^2 + 2 P_k  P_l \cos(\phi_l - \phi_k) + P_l^2  \cos(\phi_l - \phi_k)^2 + \\
& P_l^2  \sin(\phi_l - \phi_k)^2
\end{split}
\end{equation}
which, after substitution with Equation \ref{eq:square-id} again yields
\begin{equation}
P_k^2 + P_l^2 + 2 P_k P_l \cos(\phi_l - \phi_k)  = P_{kl}^2
\end{equation}
and thus
\begin{equation}
|P_k \exp(\mathrm{j} \phi_k) + P_l \exp(\mathrm{j} \phi_l)|^2 = |P_k + P_l \exp(\mathrm{j} (\phi_l - \phi_k))|^2.
\end{equation}

Therefore we can just set the phase $\phi_k$  as a reference point for our system and restructure Equation \ref{eq:state_output_relation_final} such that
\begin{equation}
\label{eq:cosine}
\cos(\phi_k - \phi_l) = \cos(\phi_k - \phi_l)= \frac{P_{kl}^2 - (P_k^2+P_l^2)}{2   P_k  P_l}.
\end{equation}
Note that, since $\phi_{kl} = \phi_l - \phi_k$ and $\phi_{lk}=\phi_k - \phi_l$ mirror each other along the real axis, their cosines are identical. Consequently, when applying an inverse cosine operation to Equation \ref{eq:cosine} we can not determine the sign of $\phi_{kl}$ but its magnitude via
\begin{equation}
\label{eq:phase-estimation-appendix}
 |\phi_{k} - \phi_{l}| = |\phi_{kl}|= \arccos\left(\frac{P_{kl}^2 - (P_k^2+P_l^2)}{2  P_k P_l}\right).
\end{equation}
We refer to Equation \ref{eq:phase-estimation-appendix} as the \emph{phase estimation equation}.

To find the actual sign of $\phi_{kl}$, additional information needs to be found. Assume we know not only $P_{kl}$, but also $P_{kl}'$ defined as
\begin{equation}
\begin{split}
P_{kl}'  =& |P_k \exp(\mathrm{j} \phi_l) \exp(\mathrm{j} \frac{\pi}{2}) + P_l \exp(\mathrm{j} \phi_l)|^2
 \\=& |P_k \exp(\mathrm{j} \phi_k') + P_l  \exp(\mathrm{j} \phi_l)|^2. 
\end{split}
\end{equation} 
Then,
\begin{equation}
\phi_{kl}'=\phi_{l} - \phi_{k} - \frac{\pi}{2}.
\end{equation}
If $\phi_{kl}$ has been mapped correctly in Equation \ref{eq:phase-estimation-appendix}, then $ 0 \leq \phi_{kl} \leq \pi$. In consequence, $(0 - \frac{\pi}{2}) \leq \phi_{kl}' \leq (\pi - \frac{\pi}{2})$, and thus will be mapped on the interval $[0,\frac{\pi}{2}]$ by the $\arccos$ function. In all remaining cases, $\phi_{kl}$ has been mapped wrong and the sign of the result of Equation \ref{eq:phase-estimation-appendix} needs to be reversed. This leaves us with a simple rule to determine the sign of $\phi_{kl}$ from $\phi_{kl}'$:
\begin{equation}
\label{eq:phase-sgn-appendix}
\phi_{st}=
   \begin{cases}
       |\phi_{kl}| & \text{if }  |\phi_{kl}'| \in [0, \frac{\pi}{2}]  \\
      -|\phi_{kl}| & \text{else}.
    \end{cases} \\
\end{equation}

\section*{Acknowledgment}
This research was funded by the EU  Horizon
2020 PHRESCO Grant (Grant No. 688579) and the BELSPO IAP
P7-35 program Photonics@be.

\ifCLASSOPTIONcaptionsoff
  \newpage
\fi

\bibliographystyle{IEEEtran}
\bibliography{IEEEabrv,nlinv}

\begin{IEEEbiography}[{\includegraphics[width=1in,height=1.25in,clip,keepaspectratio]{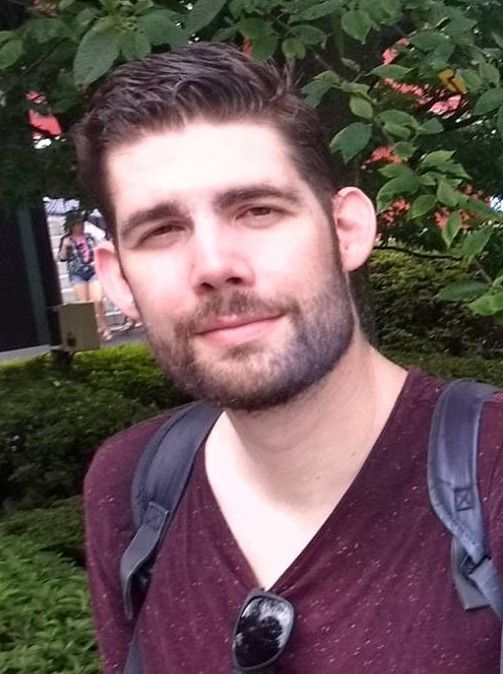}}]{Matthias Freiberger}
was born in  Graz, Austria, in 1983. He received a M.Sc. degree in  information and computer engineering  from  Graz University of Technology, Graz, Austria, in 2016. He is currently pursuing a Ph.D degree at the UGent-imec IDLab,  Department of Electronics and Information Systems at Ghent University. His current research focuses on training algorithms for neuromorphic devices. His research interests include scaling up neuromorphic systems, deep learning, and recurrent neural networks on chip.
\end{IEEEbiography}

\begin{IEEEbiography}[{\includegraphics[width=1in,height=1.25in,clip,keepaspectratio]{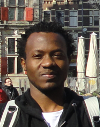}}]{Andrew Katumba}
was born in Masaka, Uganda in 1985. He received an M.Sc. degree in Optics and Photonics from Karlsrule Institute of Technology,  Germany in 2013. He is pursuing a Ph.D. degree in Photonics Engineering at the Photonics Research Group, Gent University-imec, Belgium. His current research focuses on photonic neuromorphic architectures for high speed optical telecommunications systems. He is a student member of IEEE Photonics Society and the International Society for Optics and Photonics.
\end{IEEEbiography}


\begin{IEEEbiography}[{\includegraphics[width=1in,height=1.25in,clip,keepaspectratio]{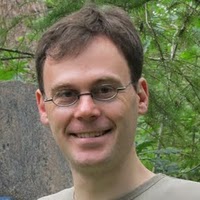}}]{Peter Bienstman}
was born in Ghent, Belgium, in 1974. He received a degree in electrical engineering from Ghent University, Belgium, in 1997 and a Ph.D. from the same university in 2001, at the Department of Information Technology (INTEC), where he is currently a full professor. His research interests include several applications of nanophotonics (biosensors, photonic information processing, ...) as well as nanophotonics modelling. He has published over 110 papers and holds several patents. He has been awarded a ERC starting grant for the Naresco-project: Novel paradigms for massively parallel nanophotonic information processing.
\end{IEEEbiography}

\begin{IEEEbiography}[{\includegraphics[width=1in,height=1.25in,clip,keepaspectratio]{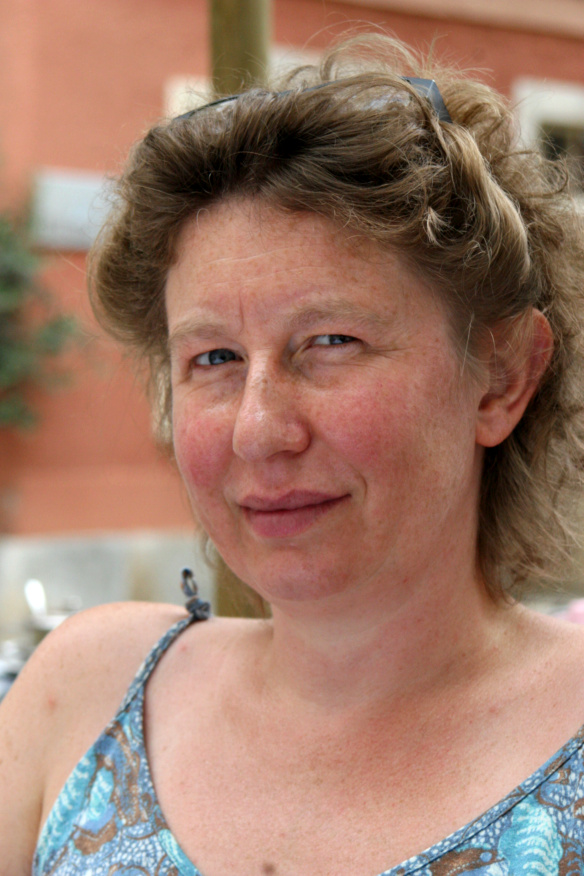}}]{Joni Dambre}
 was born in Ghent, Belgium, in 1973. She received the M.Sc. degree in electronics engineering and the Ph.D. degree in computer science engineering from Ghent University, Ghent, in 1996 and 2003, respectively. She  is currently a professor with the UGent-imec IDLab, and the Department of Electronics and Information Systems at Ghent University. She performs research on machine learning and neural networks, with applications in sensor processing and robotics, as well as innovative unconventional and neuromorphic computing hardware that exploit the computational power of physical dynamical  systems. 
\end{IEEEbiography}

\end{document}